\pdfoutput=1

\documentclass[11pt]{article}

\usepackage{EACL2023}
\usepackage{times}
\usepackage{latexsym}

\usepackage[T1]{fontenc}

\usepackage[utf8]{inputenc}

\usepackage{microtype}

\usepackage{inconsolata}
\usepackage{graphicx}
\usepackage{amsmath}
\usepackage{caption}
\usepackage{xcolor}
\usepackage{hyperref}

\title{A Quantitative Discourse Analysis of Asian Workers \\ in the US Historical Newspapers}

\author{Jaihyun Park\textsuperscript{1}, Ryan Cordell\textsuperscript{1}\\
  \textsuperscript{1}School of Information Sciences\\
  \textsuperscript{1}University of Illinois at Urbana-Champaign\\
  \texttt{jaihyun2@illinois.edu, rcordell@illinois.edu}}

\begin{document}
\maketitle
\begin{abstract}
\textcolor{red}{Warning: This paper contains examples of offensive language targetting marginalized population.}
\\
The digitization of historical texts invites researchers 
to explore the large-scale corpus of historical texts with computational methods.
In this study, we present computational text analysis on a relatively understudied topic of
how Asian workers are represented in historical newspapers in the United States. We found that the word ``coolie'' 
was semantically different in some States (e.g., Massachusetts, Rhode Island, Wyoming, Oklahoma, and Arkansas)
with the different discourses around coolie. We also found that then-Confederate newspapers and then-Union newspapers
formed distinctive discourses by measuring over-represented words. Newspapers from then-Confederate States
associated coolie with slavery-related words. In addition, we found Asians were perceived to be 
inferior to European immigrants and subjected to the target of racism. This study contributes to supplementing
the qualitative analysis of racism in the United States with quantitative discourse analysis.

\end{abstract}

\section{Introduction} \label{sec:intro}

Digitization of historical texts has opened up new opportunities for 
researchers to explore a large-scale corpus of historical texts
with computational methods. Especially, the ready availability
of datasets required for Natural Language Processing (NLP) research 
has welcomed researchers from other disciplines to NLP research \cite{park2022raison,park2023ripple} 
and reduced the barriers to entry of NLP research.
Taking this advantage, many researchers in diverse field,
namely Sociology, History, English, and Information Science 
have applied NLP techniques to historical texts, such as books \cite{parulian2022uncovering}, 
newspapers \cite{smith2013infectious,pedrazzini2022machines,santin2016or},
and/or congressional records \cite{lin2022enhancing, guldi2019parliament}
by the help of available archival metadata \cite{dobreski2019remodeling}.
Creating an interdisciplinary research space called \textit{Digital Humanities}, 
there have been studies primarily focusing on the race problem in the United States.
To introduce a few large-scale computational research, 
\citet{soni2021abolitionist} traced the semantic change of the word, for instance,
when and which newspaper started to use the word with a new meaning and when and which
newspapers adopted new semantic meaning of the word in abolitionist newspapers from the 19th century. 
\citet{franzosi2012ways} analyzed racial violence, especially lynching performed by the White mob
in the 19th century Georgia and presented quantitative narratives of the racial violence.
These studies intersect the problem of historical racism and NLP research.
However, despite the increasing attention toward the large scale text analysis on 
historical racism, there has been a gap of understanding the racism posed toward Asians in the United States.
It is true that the major racial tension in the United States has been between White and Black.
At the center of secession of the South and the creation of Confederacy, there was slavery problem. 
However, the racial tension between White and Asian has also been a part of the history of the United States.
To explore understudied topic of how Asian population was discriminated
in the history of the United States, we present computational text analysis on how Asian workers are represented in
the U.S. newspapers by searching the derogatory word (``coolie'') referencing to Asian workers. 
We further developed research questions as follows:

\begin{itemize}
  \item RQ 1. How different are the semantic meaning of ``coolie'' in each State? 
  \item RQ 2. What are the words over-represented in the newspapers between then-Confederate States and then-Union States?
  \item RQ 3. What ``coolie'' stories are reprinted and what are their characteristics?
\end{itemize}

To support open science and transparent data science, we publish the code used in this study at 
\url{https://github.com/park-jay/coolie}.

\section{Background} \label{sec:background}

Throughout 19th and 20th century, Asian immigrant workers were derogatorily called \textit{``coolies.''}
Britannica entry on \textit{``coolie''} introduces the word as ``pejorative European usage'' to 
describe ``an unskilled labourer or porter usually in or from the Far East hired for low
or subsistence wages.'' \footnote{\url{https://www.britannica.com/money/topic/coolie-Asian-labourer}}
\citet{breman1992conclusion} studied the origin of the word ``coolie''
and claimed the transformation of the word ``coolie'' from ``kuli'' 
(a type of payment for menial work in Tamil) 
signifies the change of the word from a neutral term to a derogatory term
and reflects the person collapse into the payment for labor in English.

In seeking to fill the labor shortage in the United States due to the
abolition, Chinese workers were recruited and the migration of Chinese workers arrived in the United States
beyond China-neighboring countries like India and Malaysia \cite{farley1968chinese}.
Even though Asian coolies were perceived to be 
patient, tractable, obedient, industrious, and
frugal compared to African slaves \cite{jung2006coolies}, 
coolies were distrusted, detested, and discriminated \cite{breman2023coolie}.

With influx in the number of Chinese workers in the United States,
it has come to public's attention that indentured laborers 
were analogous to modern trafficking and they are no different from
slavery \cite{kempadoo2017bound}.
\citet{jung2006coolies} argued that the United States minister to China, 
William B. Reed viewed coolie trade more than coercion and perceived the coolie problem
with the racial and national interest. Reed thought Chinese ``would either amalgamate with
with the negro race, and thus increase the actual slave population.''
Rising anti-Chinese sentiment and perceiving as a threat to White workers as their cheap replacement \cite{rhoads2002white}
and inferior, Chinese Exclusion Act of 1882 remarks the watershed of America's gatekeeping
and defining the desirability (and ``Whiteness'') of immigrant groups \cite{lee2002chinese}.

The problem of coolie exemplifies the extension of colonial and 
capitalist exploitation beyond Africa and sugarcoated 
the extended system as indentured migrant contract workers \cite{van2016coolie}.
Therefore, studying coolie problem in the United States context supplements the historical
study of racial tension mostly focused between White and Black and extends the racial conflict
to include Asian population.  

\section{Methodology} \label{sec:methodology}

\subsection{Data collection} \label{sec:data_collection}
The data is collected from Chronicling America \footnote{\url{https://chroniclingamerica.loc.gov/}} API
where digitized texts through optical character recognition (OCR) are accessible.
When the word ``coolie'' was queried, API showed 124,511 pages of newspapers containing the word \footnote{The data is collected on September 5th, 2023.}.
First, we collected the entire pages of the newspapers containing the word ``coolie.''
Then, we extracted the text from the pages and searched the exact match for the word ``coolie.'' 
This additional step ensures excluding false positive cases due to mis-recognized words.
For instance, the search included the result of ``cooli'' even though it was not identical keyword that we wanted to query.
In \textit{New York Daily Tribune} published on August 5th, 1862 contained the word ``cooli'' but it accompanied
many OCR errors making it doubtful whether the word ``cooli'' was actually from the word ``coolie.'' 
\footnote{\url{https://chroniclingamerica.loc.gov/lccn/sn83030213/1862-08-05/ed-1/seq-4/ocr/} was searched as a page that contains ``coolie''
but it was due to the OCR error.}
In order to reduce this kind of false positive cases, we double-processed the data by finding the exact match
for the word ``coolie'' in the extracted text.
Instead of finding a sentence that contains the word ``coolie'', we extracted upto ten tokens before the word ``coolie'' appeared
and upto ten tokens after the word ``coolie'' appeared to create a pseudo-sentence.
As some digitization of the newspapers were not perfect and thus missed punctuation marks, sentence tokenizer could not
identify the sentence boundary correctly.
This additional step of creating pseudo-sentence resulted in 125,253 text data (pseudo-sentence) containing the word ``coolie'' for the analysis.
The earliest publication date was June 30th, 1795 and the latest publication date was December 6th, 1963.

\begin{figure}[h!]
  \centering
  \includegraphics[width=\columnwidth]{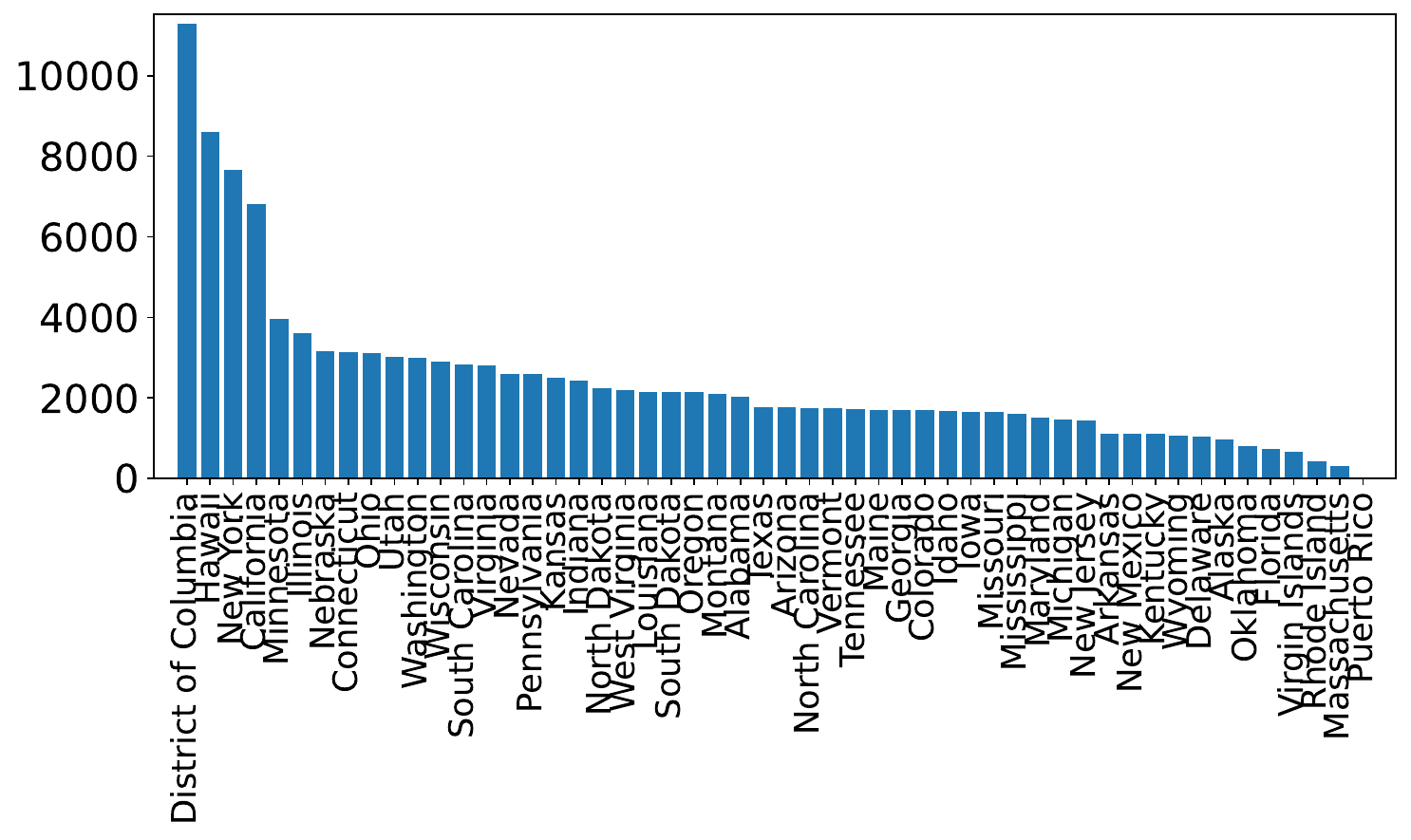}
  \caption{The count of text data containing the word ``coolie'' by State}
  \label{fig:descriptive}
\end{figure}

In figure \ref{fig:descriptive}, we present the count of text data containing the word ``coolie'' by State.
The count of text data is not evenly distributed across the States due to different digitization process of the newspapers.
The most text data was from District of Columbia (\textit{n=11,302}) followed by Hawaii (\textit{n=8,613}) and New York (\textit{n=7,671}).
Puerto Rico (\textit{n=15}), Massachusetts (\textit{n=305}), Rhode Island (\textit{n=418}), and Virgin Islands (\textit{n=656}) 
had the least text data.

\subsection{Data pre-processing} \label{sec:data_preprocessing}
We pre-processed the data by removing the punctuation, non-alphabet tokens that might have been mis-recognized
during the OCR process, and stop words. The list of stop words is from NLTK package in Python and 
we further converted the words into lemmas using Spacy \footnote{\url{https://spacy.io/}}.
Before we feed the data into the word embedding model, we ran the FastText model \cite{bojanowski_enriching_2017} 
to identify possible OCR errors. FastText embedding takes character n-gram as input and outputs the embedding vector. 
This model was tested effective that it can generate possible OCR error candidates \citep{hajiali_generating_2022}. 
By training entire sentence that contains the word ``coolie'', we identified 200 most similar words to the word ``coolie''. 
For instance, ``coolieize'' (0.8654), ``oroolie'' (0.8630), and ``roolie'' (0.8541) ranked high in the list of similar words to ``coolie''
according to the FastText embedding model.
With 200 most similar words, we changed the top 200 words in the text data into ``coolie'' and trained the Word2vec model
in section \ref{sec:method_rq1_word_embedding}. 

\subsection{RQ1. Word embedding} \label{sec:method_rq1_word_embedding}
In order to answer RQ 1, we trained the Word2vec model \cite{mikolov_distributed_2013} to
use Continuous Bags of Words (CBOW) approach and the skip-gram approach. Both CBOW and 
skip-gram approach find the word embedding by predicting the target word from the context words.
We trained the Word2vec model with minimum word count of 5 and window size of 5 to generate the word embedding. 
We then took the average of the word embedding vector of the word ``coolie'' in each State and calculated the cosine similarity.

\subsection{RQ2. Statistically over-represented words} \label{sec:method_rq2_overrepresented_words}
In answering RQ 2, we grouped the newspapers into two groups: the newspapers from the then-Confederate States and then-Union newspapers. 
For the then-Confederate States, we included the newspapers from Alabama, Arkansas, Florida, Georgia, Louisiana, Mississippi, 
North Carolina, South Carolina, Tennessee, Texas, and Virginia.
For the then-Union States, Maine, New York, New Hampshire, Vermont, Massachusetts, Connecticut, Rhode Island,
Pennsylvania, New Jersey, Ohio, Indiana, Illinois, Kansas, Michigan, Minnesota, Wisconsin, Iowa, California, Nevada, Oregon, Delaware, Maryland,
and West Virginia were included.
We excluded sentences from the newspapers newspapers located in Virgin Islands (\textit{n=656}) and Puerto Rico (\textit{n=15}).

We calculated the log-odds ratio with informative Dirichlet prior \cite{monroe2008fightin} 
by comparing the word frequency in the then-Confederacy newspapers 
and the newspapers published in the rest of the United States. 
The detailed metric is provided in equation \ref{eq: log-odds ratio}.

\begin{equation}
\begin{aligned}
  \delta_{w}^{(i-j)} = \log \frac{y_{w}^{i}+a_{w}}{n^{i}+a_{0}-y_{w}^{i}-a_{w}} \\
  - \log \frac{y_{w}^{j}+a_{w}}{n^{j}+a_{0}-y_{w}^{j}-a_{w}}
\end{aligned}
\label{eq: log-odds ratio}
\end{equation}

The log-odds ratio with informative Dirichlet of each word 
$w$ between two corpora $i$ and $j$ (in our study, newspapers from the
then-Confederate States and the rest) given the prior frequencies are obtained from the entire 
corpus $a$. We selected 15,000 most frequent words from the entire corpus and the Z-score is calculated for each word.
When $n^{i}$ is the total number of words in corpus $i$,
$y_{w}^{i}$ is the number of times word $w$ appears in corpus $i$,
$a_{0}$ is the size of the corpus $a$, and
$a_{w}$ is the frequency of word $w$ in corpus $a$ \citep{kwak2020systematic}.
With the log-odds ratio, we can identify the words that are over-represented 
in the corpora.

\subsection{RQ 3. Text reprint detection}

Nineteenth-century American newspapers reprinted texts 
from a wide range of genres: news reports, recipes, trivia, lists, vignettes, 
and religious reflections \citep{cordell2017fugitive}. 
Text reprints could also include boilerplate that appeared across many issues of the same paper, 
such as advertisements. 
A business might buy ad space for multiple weeks, months, or even years, 
and those ads would be left 
in standing type from issue to issue. In a study such as this one, focused on textual reuse,
an ad that includes a keyword of interest 
but which appears day after day can disproportionately influence the statistical 
relationship between words in the corpus, 
leading our model to overestimate the importance of words within the ad relative
to the words in texts that changed each day. 
In other words, if one particular phrase repeatedly appears, then the 
embedding model will overfit the phrase because of 
the distorted distribution of the text. However, it is hard to detect reprints based on keyword searches because of OCR errors. 
Here we adopt the Viral Texts project's text-reuse detection methods, as described in \citet{smith2014detecting} , which use n-gram document representations 
to detect text reprints 
within errorful OCR-derived text. We processed our corpus with a 5-gram chunking 
using NLTK whitespace tokenizer and further made a judgment 
that the text has been reprinted when there were more than 
three matches of 5-grams across the snippets. 
With this method, the negative impact of the OCR errors can be reduced.
For instance, this pair: [``demolish'', ``part'', ``build'', ``injure'', ``two'', ``coolie'', ``police'', ``investigation'', 
``latter'', ``case'', ``lead''] and [``demolish'', ``part'', ``build'', ``jure'', ``two'', ``latter'' 
``case'', ``lead''] are not identical because of inconsistent OCR like ``injure'' and ``jure''.
However, the 5-gram matching examination substantiated that this pair denotes a reprinted text. 
Due to the effectiveness of the method by \citet{smith2014detecting}, 
we used the method to answer RQ 3.

\section{Results} \label{sec:results}

\begin{figure*}[h!]
  \centering
  \includegraphics[width=0.65\textwidth]{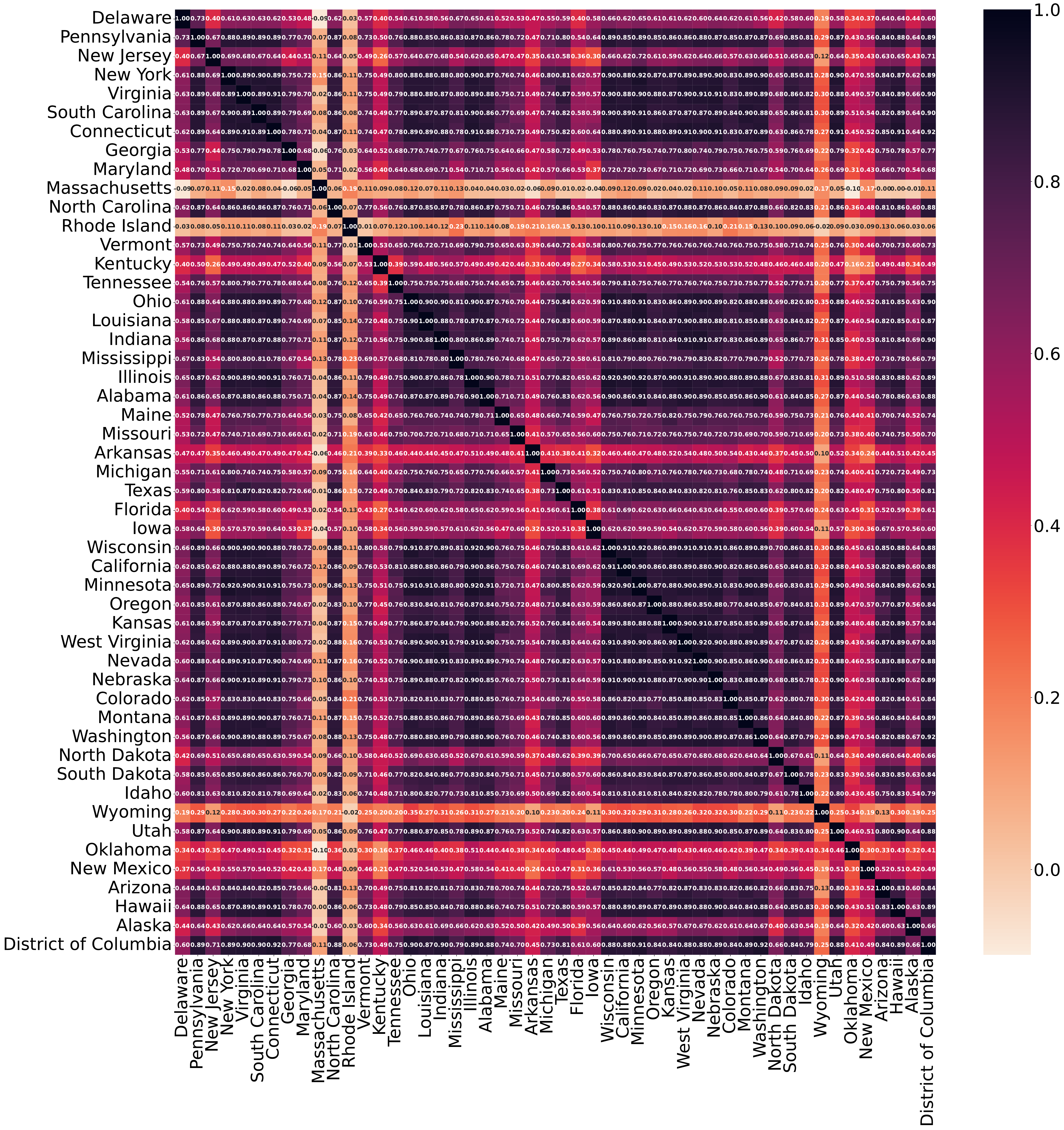}
  \caption{The heatmap of cosine similarity comparison across the average embedding vector of the word ``coolie'' in each State}
  \label{fig:heatmap}
\end{figure*}

\subsection{RQ1. Comparing the meaning of ``coolie''} \label{sec:result_rq1_word_embedding}
In figure \ref{fig:heatmap}, we present the heatmap of cosine similarity comparison across the average embedding vector of the word ``coolie'' in each State.
We are visually informed that the cosine similarity in most of the States is high except for a few States that created 
a semantically different meaning of the word ``coolie''. 
For instance, Massachusetts and Rhode Island showed average cosine similarity of 0.08 and 0.12, respectively
when average cosine similarity across the entire States was 0.65.
The most semantically dissimilar State to Massachusetts was when compared to Oklahoma (-0.10) and the most similar State to Massachusetts was North Dakota (0.23). 
In the meantime, Rhode Island showed the most dissimilar meaning of the word ``coolie'' when compared to Delaware (-0.03) and the most similar State to Rhode Island was Mississippi (0.23). 
Some then-Confederate States showed lower cosine similarity than the average cosine similarity across the entire States.
For instance, Arkansas (0.43), Florida (0.48), and Tennessee (0.63) showed lower cosine similarity on average. 
The most dissimilar State to Arkansas was Massachusetts (-0.06) and the most similar State to Arkansas was Colorado (0.54).
For Florida, the most dissimilar State was Rhode Island (0.06) and the most similar State to Florida was Nevada and Utah (0.61).
For Tennessee, the most dissimilar State was Massachusetts (-0.03) and the most similar State to Tennessee was Utah (0.80).

\begin{table}[h!]
  \resizebox{\columnwidth}{!}{%
  \begin{tabular}{|ccccc|}
  \hline
  \multicolumn{5}{|c|}{The highest five States} \\ \hline
  \multicolumn{1}{|c|}{Illinois} & \multicolumn{1}{c|}{California} & \multicolumn{1}{c|}{Wisconsin} & \multicolumn{1}{c|}{Virginia} & Nevada \\ \hline
  \multicolumn{1}{|c|}{\begin{tabular}[c]{@{}c@{}}labor\\ (0.9998)\end{tabular}} & \multicolumn{1}{c|}{\begin{tabular}[c]{@{}c@{}}country\\ (0.9995)\end{tabular}} & \multicolumn{1}{c|}{\begin{tabular}[c]{@{}c@{}}chinese\\ (0.9998)\end{tabular}} & \multicolumn{1}{c|}{\begin{tabular}[c]{@{}c@{}}chinese\\ (0.9998)\end{tabular}} & \begin{tabular}[c]{@{}c@{}}chinese\\ (0.9997)\end{tabular} \\ \hline
  \multicolumn{1}{|c|}{\begin{tabular}[c]{@{}c@{}}chinese\\ (0.9998)\end{tabular}} & \multicolumn{1}{c|}{\begin{tabular}[c]{@{}c@{}}bill\\ (0.9995)\end{tabular}} & \multicolumn{1}{c|}{\begin{tabular}[c]{@{}c@{}}labor\\ (0.9998)\end{tabular}} & \multicolumn{1}{c|}{\begin{tabular}[c]{@{}c@{}}man\\ (0.9997)\end{tabular}} & \begin{tabular}[c]{@{}c@{}}club\\ (0.9997)\end{tabular} \\ \hline
  \multicolumn{1}{|c|}{\begin{tabular}[c]{@{}c@{}}wage\\ (0.9997)\end{tabular}} & \multicolumn{1}{c|}{\begin{tabular}[c]{@{}c@{}}upon\\ (0.9995)\end{tabular}} & \multicolumn{1}{c|}{\begin{tabular}[c]{@{}c@{}}two\\ (0.9998)\end{tabular}} & \multicolumn{1}{c|}{\begin{tabular}[c]{@{}c@{}}trade\\ (0.9997)\end{tabular}} & \begin{tabular}[c]{@{}c@{}}labor\\ (0.9997)\end{tabular} \\ \hline
  \multicolumn{1}{|c|}{\begin{tabular}[c]{@{}c@{}}two\\ (0.9997)\end{tabular}} & \multicolumn{1}{c|}{\begin{tabular}[c]{@{}c@{}}well\\ (0.9995)\end{tabular}} & \multicolumn{1}{c|}{\begin{tabular}[c]{@{}c@{}}one\\ (0.9998)\end{tabular}} & \multicolumn{1}{c|}{\begin{tabular}[c]{@{}c@{}}work\\ (0.9997)\end{tabular}} & \begin{tabular}[c]{@{}c@{}}make\\ (0.9997)\end{tabular} \\ \hline
  \multicolumn{1}{|c|}{\begin{tabular}[c]{@{}c@{}}one\\ (0.9997)\end{tabular}} & \multicolumn{1}{c|}{\begin{tabular}[c]{@{}c@{}}go\\ (0.9995)\end{tabular}} & \multicolumn{1}{c|}{\begin{tabular}[c]{@{}c@{}}time\\ (0.9998)\end{tabular}} & \multicolumn{1}{c|}{\begin{tabular}[c]{@{}c@{}}one\\ (0.9997)\end{tabular}} & \begin{tabular}[c]{@{}c@{}}use\\ (0.9996)\end{tabular} \\ \hline
  \multicolumn{1}{|c|}{\begin{tabular}[c]{@{}c@{}}day\\ (0.9997)\end{tabular}} & \multicolumn{1}{c|}{\begin{tabular}[c]{@{}c@{}}stop\\ (0.9995)\end{tabular}} & \multicolumn{1}{c|}{\begin{tabular}[c]{@{}c@{}}carry\\ (0.9997)\end{tabular}} & \multicolumn{1}{c|}{\begin{tabular}[c]{@{}c@{}}three\\ (0.9997)\end{tabular}} & \begin{tabular}[c]{@{}c@{}}say\\ (0.9996)\end{tabular} \\ \hline
  \multicolumn{1}{|c|}{\begin{tabular}[c]{@{}c@{}}china\\ (0.9997)\end{tabular}} & \multicolumn{1}{c|}{\begin{tabular}[c]{@{}c@{}}many\\ (0.9995)\end{tabular}} & \multicolumn{1}{c|}{\begin{tabular}[c]{@{}c@{}}take\\ (0.9997)\end{tabular}} & \multicolumn{1}{c|}{\begin{tabular}[c]{@{}c@{}}number\\ (0.9997)\end{tabular}} & \begin{tabular}[c]{@{}c@{}}importation\\ (0.9996)\end{tabular} \\ \hline
  \multicolumn{1}{|c|}{\begin{tabular}[c]{@{}c@{}}man\\ (0.9997)\end{tabular}} & \multicolumn{1}{c|}{\begin{tabular}[c]{@{}c@{}}american\\ (0.9995)\end{tabular}} & \multicolumn{1}{c|}{\begin{tabular}[c]{@{}c@{}}make\\ (0.9997)\end{tabular}} & \multicolumn{1}{c|}{\begin{tabular}[c]{@{}c@{}}make\\ (0.9997)\end{tabular}} & \begin{tabular}[c]{@{}c@{}}man\\ (0.9996)\end{tabular} \\ \hline
  \multicolumn{1}{|c|}{\begin{tabular}[c]{@{}c@{}}pay\\ (0.9997)\end{tabular}} & \multicolumn{1}{c|}{\begin{tabular}[c]{@{}c@{}}con\\ (0.9995)\end{tabular}} & \multicolumn{1}{c|}{\begin{tabular}[c]{@{}c@{}}japanese\\ (0.9997)\end{tabular}} & \multicolumn{1}{c|}{\begin{tabular}[c]{@{}c@{}}importation\\ (0.9997)\end{tabular}} & \begin{tabular}[c]{@{}c@{}}trade\\ (0.9996)\end{tabular} \\ \hline
  \multicolumn{1}{|c|}{\begin{tabular}[c]{@{}c@{}}say\\ (0.9997)\end{tabular}} & \multicolumn{1}{c|}{\begin{tabular}[c]{@{}c@{}}would\\ (0.9995)\end{tabular}} & \multicolumn{1}{c|}{\begin{tabular}[c]{@{}c@{}}would\\ (0.9997)\end{tabular}} & \multicolumn{1}{c|}{\begin{tabular}[c]{@{}c@{}}two\\ (0.9997)\end{tabular}} & \begin{tabular}[c]{@{}c@{}}day\\ (0.9996)\end{tabular} \\ \hline
  \end{tabular}%
  }
  \caption{Top 10 most similar words to the word ``coolie'' in the top 5 States that showed the most similar meaning of the word ``coolie''}
  \label{tab:top10_similar_words}
  \end{table}

To delve into the semantic difference of the word ``coolie'' in each State, we present the top 10 most similar words to the word ``coolie'' in 
the top 5 States that showed the most similar meaning of the word ``coolie'' and the top 5 States that showed the most dissimilar meaning of the word ``coolie'' 
in table \ref{tab:top10_similar_words} and table \ref{tab:bottom10_similar_words}.

In table \ref{tab:top10_similar_words}, we can observe that the word ``coolie'' was used in the context of labor, China, and wage.
The word related to labor (``labor'' and ``work'') appeared in the semantically close words in Illinois, Wisconsin, Virginia, and Nevada.
``chinese'' was the most similar word of identification of coolie's ethnicity (Illinois, Wisconsin, Virginia, and Nevada) while
``japanese'' appeared in Wisconsin as well. 
The word related to wage (``wage'' and ``pay'') appeared in the high closest word to ``coolie.''

Among the highest five States, there are many common words across the States that might have created an embedding for the word ``coolie''
not so much different from other States. Words like ``make'', ``say'', ``man'', and/or numbers like ``one'' and ``two'' are common words
across the highest five States.

\begin{table}[h!]
  \resizebox{\columnwidth}{!}{%
  \begin{tabular}{|ccccc|}
  \hline
  \multicolumn{5}{|c|}{The lowest five States} \\ \hline
  \multicolumn{1}{|c|}{Massachusetts} & \multicolumn{1}{c|}{Rhode Island} & \multicolumn{1}{c|}{Wyoming} & \multicolumn{1}{c|}{Oklahoma} & Arkansas \\ \hline
  \multicolumn{1}{|c|}{\begin{tabular}[c]{@{}c@{}}among\\ (0.3579)\end{tabular}} & \multicolumn{1}{c|}{\begin{tabular}[c]{@{}c@{}}order\\ (0.4116)\end{tabular}} & \multicolumn{1}{c|}{\begin{tabular}[c]{@{}c@{}}chinese\\ (0.9969)\end{tabular}} & \multicolumn{1}{c|}{\begin{tabular}[c]{@{}c@{}}chinese\\ (0.9821)\end{tabular}} & \begin{tabular}[c]{@{}c@{}}labor\\ (0.9985)\end{tabular} \\ \hline
  \multicolumn{1}{|c|}{\begin{tabular}[c]{@{}c@{}}call\\ (0.3201)\end{tabular}} & \multicolumn{1}{c|}{\begin{tabular}[c]{@{}c@{}}india\\ (0.3972)\end{tabular}} & \multicolumn{1}{c|}{\begin{tabular}[c]{@{}c@{}}labor\\ (0.9969)\end{tabular}} & \multicolumn{1}{c|}{\begin{tabular}[c]{@{}c@{}}shoulder\\ (0.9815)\end{tabular}} & \begin{tabular}[c]{@{}c@{}}chinese\\ (0.9984)\end{tabular} \\ \hline
  \multicolumn{1}{|c|}{\begin{tabular}[c]{@{}c@{}}know\\ (0.2832)\end{tabular}} & \multicolumn{1}{c|}{\begin{tabular}[c]{@{}c@{}}woman\\ (0.3922)\end{tabular}} & \multicolumn{1}{c|}{\begin{tabular}[c]{@{}c@{}}would\\ (0.9955)\end{tabular}} & \multicolumn{1}{c|}{\begin{tabular}[c]{@{}c@{}}japanese\\ (0.9776)\end{tabular}} & \begin{tabular}[c]{@{}c@{}}mongolian\\ (0.9978)\end{tabular} \\ \hline
  \multicolumn{1}{|c|}{\begin{tabular}[c]{@{}c@{}}prohibit\\ (0.2295)\end{tabular}} & \multicolumn{1}{c|}{\begin{tabular}[c]{@{}c@{}}great\\ (0.3890)\end{tabular}} & \multicolumn{1}{c|}{\begin{tabular}[c]{@{}c@{}}japanese\\ (0.9952)\end{tabular}} & \multicolumn{1}{c|}{\begin{tabular}[c]{@{}c@{}}pay\\ (0.9748)\end{tabular}} & \begin{tabular}[c]{@{}c@{}}one\\ (0.9978)\end{tabular} \\ \hline
  \multicolumn{1}{|c|}{\begin{tabular}[c]{@{}c@{}}time\\ (0.2232)\end{tabular}} & \multicolumn{1}{c|}{\begin{tabular}[c]{@{}c@{}}take\\ (0.3761)\end{tabular}} & \multicolumn{1}{c|}{\begin{tabular}[c]{@{}c@{}}six\\ (0.9950)\end{tabular}} & \multicolumn{1}{c|}{\begin{tabular}[c]{@{}c@{}}also\\ (0.9725)\end{tabular}} & \begin{tabular}[c]{@{}c@{}}thousand\\ (0.9977)\end{tabular} \\ \hline
  \multicolumn{1}{|c|}{\begin{tabular}[c]{@{}c@{}}report\\ (0.2221)\end{tabular}} & \multicolumn{1}{c|}{\begin{tabular}[c]{@{}c@{}}ship\\ (0.3745)\end{tabular}} & \multicolumn{1}{c|}{\begin{tabular}[c]{@{}c@{}}one\\ (0.9949)\end{tabular}} & \multicolumn{1}{c|}{\begin{tabular}[c]{@{}c@{}}labor\\ (0.9706)\end{tabular}} & \begin{tabular}[c]{@{}c@{}}japanese\\ (0.9975)\end{tabular} \\ \hline
  \multicolumn{1}{|c|}{\begin{tabular}[c]{@{}c@{}}get\\ (0.2209)\end{tabular}} & \multicolumn{1}{c|}{\begin{tabular}[c]{@{}c@{}}law\\ (0.3432)\end{tabular}} & \multicolumn{1}{c|}{\begin{tabular}[c]{@{}c@{}}bring\\ (0.9948)\end{tabular}} & \multicolumn{1}{c|}{\begin{tabular}[c]{@{}c@{}}carry\\ (0.9632)\end{tabular}} & \begin{tabular}[c]{@{}c@{}}tolerate\\ (0.9975)\end{tabular} \\ \hline
  \multicolumn{1}{|c|}{\begin{tabular}[c]{@{}c@{}}arrive\\ (0.2176)\end{tabular}} & \multicolumn{1}{c|}{\begin{tabular}[c]{@{}c@{}}united\\ (0.3260)\end{tabular}} & \multicolumn{1}{c|}{\begin{tabular}[c]{@{}c@{}}say\\ (0.9946)\end{tabular}} & \multicolumn{1}{c|}{\begin{tabular}[c]{@{}c@{}}home\\ (0.9632)\end{tabular}} & \begin{tabular}[c]{@{}c@{}}revivial\\ (0.9974)\end{tabular} \\ \hline
  \multicolumn{1}{|c|}{\begin{tabular}[c]{@{}c@{}}come\\ (0.2131)\end{tabular}} & \multicolumn{1}{c|}{\begin{tabular}[c]{@{}c@{}}carry\\ (0.3234)\end{tabular}} & \multicolumn{1}{c|}{\begin{tabular}[c]{@{}c@{}}work\\ (0.9944)\end{tabular}} & \multicolumn{1}{c|}{\begin{tabular}[c]{@{}c@{}}get\\ (0.9630)\end{tabular}} & \begin{tabular}[c]{@{}c@{}}may\\ (0.9974)\end{tabular} \\ \hline
  \multicolumn{1}{|c|}{\begin{tabular}[c]{@{}c@{}}two\\ (0.2104)\end{tabular}} & \multicolumn{1}{c|}{\begin{tabular}[c]{@{}c@{}}many\\ (0.3122)\end{tabular}} & \multicolumn{1}{c|}{\begin{tabular}[c]{@{}c@{}}two\\ (0.9941)\end{tabular}} & \multicolumn{1}{c|}{\begin{tabular}[c]{@{}c@{}}work\\ (0.9612)\end{tabular}} & \begin{tabular}[c]{@{}c@{}}carry\\ (0.9974)\end{tabular} \\ \hline
  \end{tabular}%
  }
  \caption{Top 10 most similar words to the word ``coolie'' in the top 5 States that showed the most dissimilar meaning of the word ``coolie''}
  \label{tab:bottom10_similar_words}
  \end{table}

  \begin{figure*}[h!]
    \centering
    \includegraphics[width=0.65\textwidth]{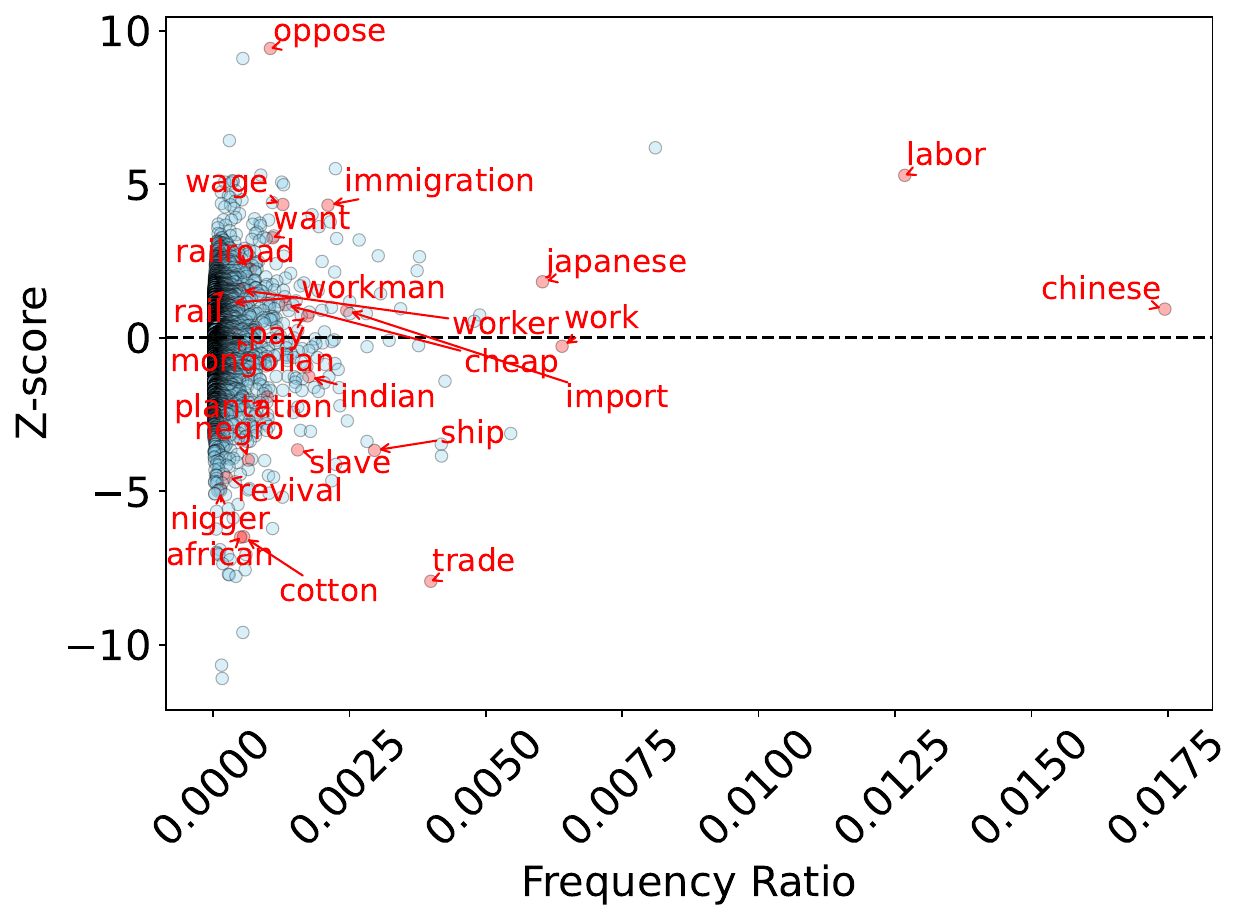}
    \caption{The Z-score of words in then-Confederate and then-Union newspapers}
    \label{fig:zscore}
  \end{figure*}

However, the most dissimilar States, Massachusetts and Rhode Island, showed the different context of the word ``coolie''
as presented in table \ref{tab:bottom10_similar_words}. Unlike the highest five States, where the word ``coolie'' was used in the context of
labor, China, and wage, Massachusetts and Rhode Island created a unique discourse of the word ``coolie.''
For instance, the word ``labor'' and ``work'' does not appear in the top 10 most similar words to the word ``coolie'' in Massachusetts and Rhode Island.
Similarly, we cannot find ``chinese'' or ``japanese'' that refer to the ethnicity of coolie workers in the top 10 most similar words as well as ``wage'' or ``pay.''
Instead, Massachusetts has ``among'', ``call'', ``know'', ``prohibit'', ``report'', ``get'', ``arrive'', and ``come'' as
semantically close words to ``coolie.'' However, these words are not found in top 10 closest words in the highest five States in table \ref{tab:top10_similar_words}.
A unique embedding for the word ``coolie'' in Rhode Island could also be attributed to the unique discourse created by uncommon words
such as ``order'', ``india'', ``woman'', ``great'', ``ship'', ``law'', ``united'', and ``many.''
However, Wyoming, Oklahoma, and Arkansas share common words with the States in table \ref{tab:top10_similar_words}. 
The words like ``chinese'', ``labor'', ``would'', ``japanese'', ``one'', ``say'', ``work'' in Wyoming are among top 10 similar words in
the highest five States. Despite existence of common words, ``shoulder'', and ``home'' in Oklahoma and ``mongolian'', ``tolerate'', and ``revival''
in Arkansas are unique words that are not found in the top 10 similar words in the highest five States.

\subsection{RQ 2. Statistically over-represented words in then-Confederate and then-Union States} \label{sec:result_rq2_overrepresented_words}

We present the result of the log-odds ratio with informative Dirichlet prior in figure \ref{fig:zscore}.
The words that are over-represented in the then-Confederate States are located in the area of below 0 while 
the words that are over-represented in the then-Union States are located in the area of above 0. X-axis represents
the frequency ratio of the word in both then-Confederate and then-Union States. Y-axis represents the Z-score of the word.
The most frequent word ``chinese'' have 1.0681 Z-score meaning that the word ``chinese'' is used relatively
more frequently in the then-Union States newspapers than the then-Confederate States newspapers.
For ``labor'', it was over-represented in the then-Union States newspapers with 5.7346 Z-score while
semantically similar word ``work'' was a little skewed to the then-Union States newspapers with -0.5145 Z-score. 
On the contrary, ``worker'' and ``workman'' are over-represented in the then-Union States newspapers with
1.7586 and 1.1122 Z-score, respectively.
The word concerning compensation for indentured labor, such as the word ``wage'' (Z-score=4.3827), \
``cheap'' (Z-score=1.2223), and ``pay'' (Z-score=0.5588) are over-represented in the then-Union States newspapers.
The word about coolie trade (``trade'' and ``ship'') are over-represented in the then-Confederate States newspapers with
Z-score of -8.0541 and -4.1275 respectively.
The word related to the race whose labor was exploited under the slavery institution, such as 
``slave'' (Z-score=-3.704), ``negro'' (Z-score=-4.1616), ``nigger'' (Z-score=-4.8824), and ``african'' (Z-score=-6.4188) are largely over-represented 
in the then-Confederate States newspapers. In addition, the location where the labor force was most wanted from coolies
were well-captured by log-odds ratio metric. For instance, ``plantation'' (Z-score=-2.0234) and 
``cotton'' (Z-score=-6.4234) are over-represented
in the then-Confederate newspapers while ``rail'' (Z-score=1.7115) and ``railroad'' (Z-score=2.3166) are over-represented in the then-Union newspapers.
The finding that the discourse around coolie is associated with slavery-related words in the then-Confederate newspapers 
supplements historical claim that the indentured Asian laborers were introduced to fill the absence of labor force in the South after abolition
of free labor of African Americans \cite{van2016coolie}.

\subsection{RQ 3. Reprint network of coolie stories} \label{sec:result_rq3_reprint}

\begin{figure}[h!]
  \centering
  \includegraphics[width=0.65\textwidth]{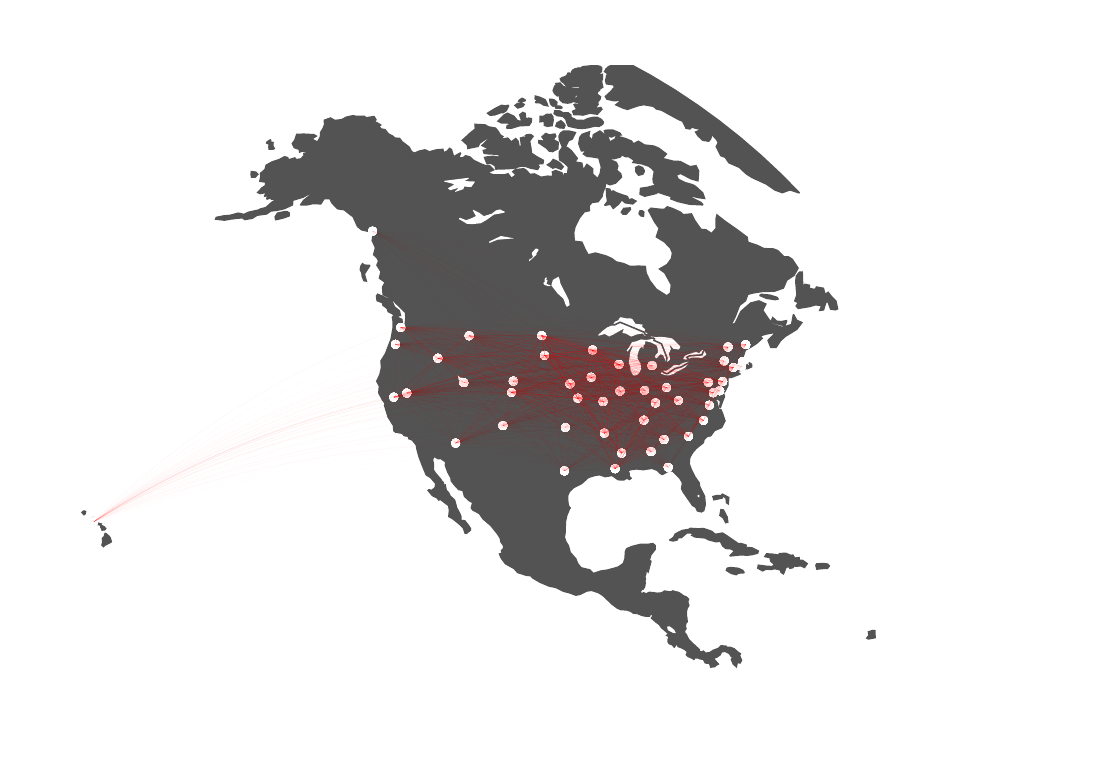}
  \caption{The reprint network of ``coolie'' stories in the newspapers}
  \label{fig:reprint}
\end{figure}

In this section, we present the network of coolie story reprints in figure \ref{fig:reprint}. 
We represented the Capital city of each State instead of connecting the cities where the newspaper company was located 
for the sake of brevity of visualization.
The network of text reprint about coolie stories shows 
high average clustering coefficient (0.9905) \cite{saramaki2007generalizations} due to 
the presence of reprints spread to multiple States.
We found the most reprinted text is from the Democratic party
at National Convention. The political message contained the 
word ``coolie.'' We found 97 reprints of the text from declaration
speech from \textit{The Opelousas courier} published in July 8th, 1876.

\begin{figure}[h!]
  \centering
  \includegraphics[width=\columnwidth]{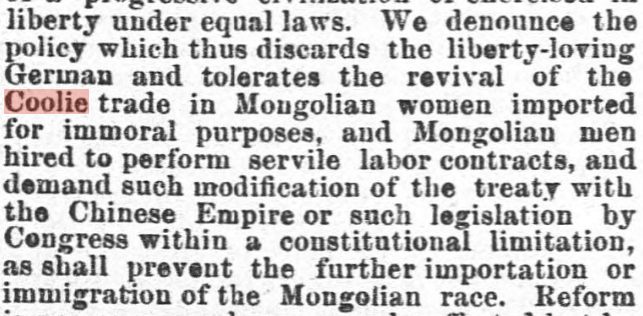}
  \caption{The text containing ``coolie'' in \textit{The Opelousas courier} published on July 8th, 1876}
  \label{fig:excerpt1}
\end{figure}

The excerpt of the message from the Democratic party is presented in figure \ref{fig:excerpt1}.
This Statement also implies that Asians (represented as Mongolian with derogatory term) are
inferior to ``liberty-loving'' Germans who could have spread the idea of freedom and solidified
the spirit of liberty in the United States that coolies could not bring. 
Another comparison can be made with treating Asian workers as commodities instead of
human beings by calling ``coolie trade'' and ``importation.''
Indeed, the banning of coolie transportation is based on viewing it as 
illegal, immoral, and inhuman atrocities resembling African slave trade \cite{jung2006coolies}.
However, coolies are unwelcoming race compared to European immigrants.
Supplementing \citet{lee2002chinese}'s argument that the desirable quality of 
immigrants entering the United States was racially White, we observe that there was a prevailing discourse
of discriminating Asians against so-called ``liberty-loving'' Germans. 

\begin{figure}[h!]
  \centering
  \includegraphics[width=0.70\columnwidth]{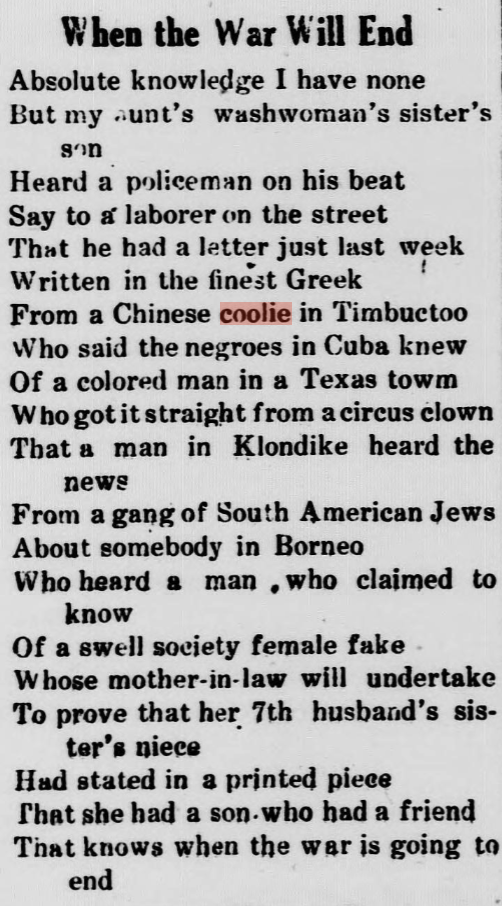}
  \caption{The text containing ``coolie'' in \textit{Middletown transcript} published on April 13th, 1918}
  \label{fig:excerpt2}
\end{figure}

The next common reprint in the dataset was from the poem that was circulated through 
78 reprints. The poem is presented in figure \ref{fig:excerpt2} and it was
appeared in \textit{Middletown transcript} published in April 13th, 1918.
The overall sentiment in this poem is light and jokey, however, this poem exemplifies the history
when racism was naturalized in cultural discourse. 
This poem borrows marginalized population such as ``coolie,'' ``negroes,'' ``colored man,''
and ``jews'' which are not necessarily critical components that
make rhymes work. Although this poem reflects orientalist view \cite{orientalism1978} by bringing
``Timbuctoo,'' ``Greek,'' and ``Klondike '' and putting emphasis on exotic sense of information about the war,
it can be offensive to people who are settling in the United States from the places
mentioned in the poem. The poem emphasizes the distance and exoticism of the places
and people who are not White and thus it can be considered as micro aggression toward
marginally represented population in the United States.



\section{Conclusion} \label{sec:conclusion}
In this study, we present a quantitative discourse analysis on ``coolie,'' 
a derogatory term referencing to Asian workers
which has been understudied in digital humanities field.
We used word embedding to compare the meaning of ``coolie'' in each State
and found that Massachusetts, Rhode Island, Wyoming, Oklahoma, and Arkansas
showed the most dissimilar meaning of the word ``coolie'' while Illinois, California, Wisconsin, Virginia, and Nevada
showed the most similar meaning of the word ``coolie'' compared to the rest of the States. 
We found the reason for the dissimilar meaning of the word ``coolie'' in Massachusetts and Rhode Island
could be in part due to the unique discourse created by uncommon words. 
With log-odds ratio calculation, we found that the discourse of coolie in the then-Confederate newspapers
was accompanied by the words related to African American slavery as well as where the 
work force is the most needed (e.g., ``cotton'' and ``plantation''). This finding supplements
historical argument that Asian workers were introduced to the United States to replace African American labor.
With text reuse detection, we found discriminating expression toward Asian workers 
in political Statement and poems that could show stereotype of Asian workers in the United States history. 

\section{Limitations and Future Work} \label{sec:limitations}
In addition to data-inherited limitation, OCR errors, 
digital archives are far from objective and neutral reflection of the past. 
Digitization of cultural and historical materials is influenced by power politics 
and it requires a caution in interpreting the results \cite{zaagsma2023digital} as the result might be grounded
in skewed number of available data. 
For instance, Chronicling America has been criticized for
imbalanced number of available data, presenting dominant viewpoints of White compared 
to small number of digitized Black press \cite{fagan2016chronicling}. 
Massachusetts and Rhode Island, the States that showed the most dissimilar meaning of the word ``coolie''
in our study, are the States that have relatively few number of newspapers containing the word ``coolie.''
As number of available digitized data does not reflect the actual number of newspapers published in the past,
we need to be cautious in interpreting the results.
In addition, the boundary of the State is not fixed and it is not clear whether the State boundary
in the past is the same as the State boundary in the present. 
With gold rush and railroad construction, the West was rapidly developed and the State boundary
was changed. Therefore, after the Civil War, grouping the States into the then-Confederate and then-Union
States might not be the best way to group the States as it does not reflect the Westward expansion of the United States.
In future work, we will explore more subtle unit of analysis such as geographical location after 
the civil war to include the rise of the West and how introduction of Chinese Exclusion Act of 1882
changed the meaning of coolie. 

\bibliography{anthology,custom}
\bibliographystyle{acl_natbib}



\end{document}